\documentclass{article}
\usepackage{ijcai23}

\usepackage{times}
\usepackage{soul}
\usepackage{url}
\usepackage[hidelinks]{hyperref}
\usepackage[utf8]{inputenc}
\usepackage[small]{caption}
\usepackage{graphicx}
\usepackage{amsmath}
\usepackage{amsthm}
\usepackage{booktabs}
\usepackage{algorithm}
\usepackage{algorithmic}
\usepackage[switch]{lineno}
\usepackage{amssymb}
\usepackage{subfigure}
\usepackage{newfloat}
\usepackage{listings}

\usepackage{amsfonts}
\usepackage{multirow}
\usepackage{color} 

\DeclareMathOperator{\subt}{s.t.}

\title{Style Miner: Find Significant and Stable Explanatory Factors in Time Series with Constrained Reinforcement Learning}

\author{
Dapeng Li$^1$
\and
Feiyang Pan$^2$\and
Jia He$^{2}$\and
Zhiwei Xu$^{1}$\and
Dandan Tu$^{2}$\And
Guoliang Fan$^1$
\affiliations
$^1$Institute of Automation,Chinese Academy of Sciences\\
$^2$Huawei EI Innovation Lab\\
\emails
lidapeng2020@ia.ac.cn,
pfy824@gmail.com}

\begin{document}

\maketitle

\begin{abstract}
In high-dimensional time-series analysis, it is essential to have a set of key factors (namely, the \emph{style factors}) that explain the change of the observed variable. For example, volatility modeling in finance relies on a set of risk factors, and climate change studies in climatology rely on a set of causal factors. The ideal low-dimensional style factors should balance significance (with high explanatory power) and stability (consistent, no significant fluctuations). However, previous supervised and unsupervised feature extraction methods can hardly address the tradeoff. In this paper, we propose Style Miner, a reinforcement learning method to generate style factors. We first formulate the problem as a Constrained Markov Decision Process with explanatory power as the return and stability as the constraint. Then, we design fine-grained immediate rewards and costs and use a Lagrangian heuristic to balance them adaptively. Experiments on real-world financial data sets show that Style Miner outperforms existing learning-based methods by a large margin and achieves a relatively 10\% gain in
R-squared explanatory power compared to the industry-renowned factors proposed by human experts.
\end{abstract}
\section{Introduction}

\begin{figure}[t] 
    \centering
    \includegraphics[width=0.45\textwidth]{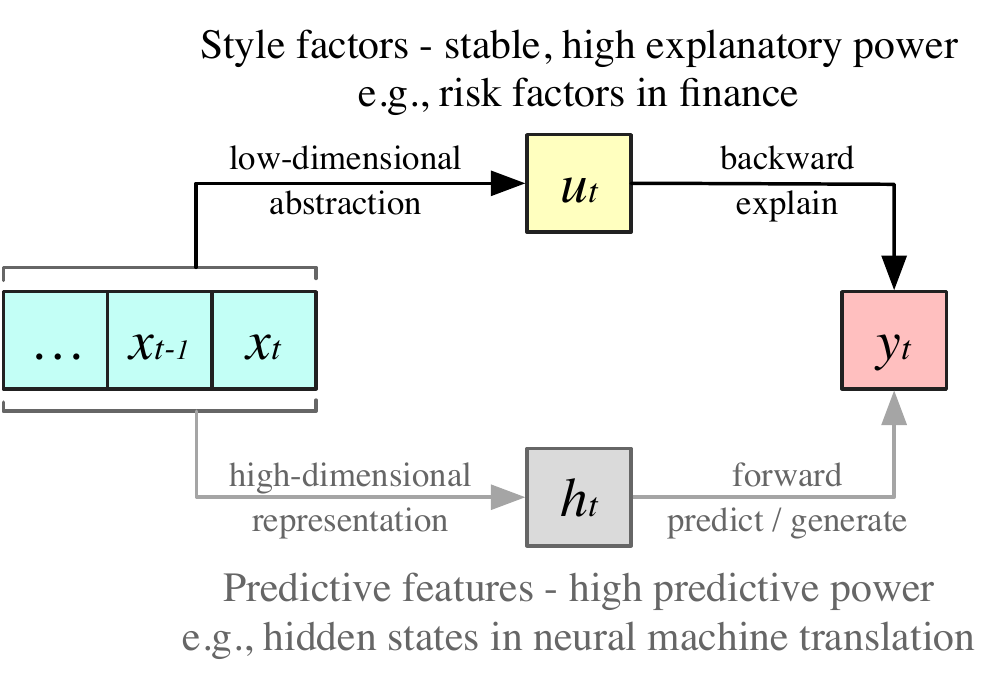}
    \caption{Comparison between style and predictive factors.}
    \label{fig1:StyleFactor}
\end{figure}

As is known to all, feature extraction is crucial for data analysis and machine learning in the real world. Generally speaking, features have two major functions: for prediction and for explanation. Nowadays, predictive features can be extracted efficiently with a great number of tools in the supervised machine learning toolbox, and made a great deal of progress in many areas, including computer vision~\cite{34:ping2013review}, neural machine translation~\cite{35:bahdanau2014neural}, and so on. However, there still remain some real-world problems where the future observations are hard to predict, such as the equity market and climate change. In these circumstances, it is more important to find the factors that have strong explanatory power so that the complex real-world states can be abstracted and thus can be understood by people. For instance, economists~\cite{21:ilmanen2012understanding} use low-dimensional style factors to explain asset volatility to understand the market, measure asset risks, and establish portfolios. Meteorologists~\cite{23:hegerl2011use,25:bindoff2013detection} use greenhouse gas, other anthropogenic stressors, and natural components to explain the observed global surface temperature changes so the global warming can be mitigated by controlling the main style factors. In this paper, we would like to answer the question of how to use reinforcement learning to find explanatory style factors especially in high-dimensional time-series problems, which explain the observations rather than predict them, as illustrated in Figure~\ref{fig1:StyleFactor}.
 
As discussed in~\citeauthor{28:menchero2011barra}[\citeyear{28:menchero2011barra}], high-quality style factors should meet two basic requirements: have high explanatory power, and be stable. First, the main usage of styles is to explain the future outcomes, and the explanation should be significant. Second, styles must be stable (have little fluctuation) so that historically lagged explanations can be valuable for the future decisions. In this paper, we focus on finding style factors that meet both requirements.


There are three lines of studies for style factor extraction, which are 1) case-by-case studies by human experts, 2) unsupervised feature extraction methods, and 3) supervised feature extraction methods. The design of expert factors \cite{2:1975The,3:sheikh1996barra,4:sharpe1964capital} usually requires strong expertise in specific areas and can achieve reasonable performance. However, in the area of high-dimensional big data, finding new style factors in novel datasets gets harder and harder. Unsupervised methods such as principal components analysis (PCA)~\cite{5:wold1987principal} and AutoEncoder~\cite{13:gu2021autoencoder} can find hidden variables in static datasets. However, they do not show promising performance in time-series data when the purpose is to explain unobserved future outcomes rather than current observations. Supervised learning methods, on the other hand, can handle sequential datasets with the recent progress of sequential neural networks. For example, Deep Risk Model (DRM)~\cite{7:lin2021deep} is shown to perform better than expert factors. However, as supervised deep learning methods rely on differentiable loss functions, they are hard to apply to problems with non-differentiable feedback. 
There are also some studies using Reinforcement Learning method in other finance domain~\cite{lcad}, however, it cannot be directly applied to style factor mining.

In this paper, we propose Style Miner, a style factor extraction method based on constrained reinforcement learning which explicitly addresses the tradeoff between explanatory power and stability. We first formulate the style factor extraction as a Constrained Markov Decision Process (CMDP)~\cite{32:altman1999constrained}, where an agent tries to generate continuous style factors given the time-series observations. For each generated style factor, the agent receives the explanatory power as the long-term return and the negative autocorrelation of the factor sequence as the cost. The agent's goal is to maximize the explanatory power while satisfying the constraint of having a high sequence autocorrelation. Further, to accelerate the training, we also propose several practical techniques to simplify the problem and an adaptive heuristic to explicitly balance the reward and the constraint. Experiments in real-world financial market data show that Style Miner can achieve state-of-the-art performance while being stable given only the raw market data inputs, which achieves a relatively 10\% higher R-squared explanatory power compared to expert factors.

The main contributions of this thesis are summarized as follows:
\begin{itemize}
    \item We propose Style Miner which extracts style factors with Constrained Reinforcement Learning. It tries to maximize the explanatory power while constraining the stability. To the best of our knowledge, it is the first study that use reinforcement learning method to extract style factors.
    \item We design fine-grained immediate rewards and costs to alleviate the sparse reward problem, and then use a Lagrangian heuristic to adaptively balance between the reward and the cost constraint.
    \item We conduct extensive experiments and show that our method can achieve state-of-the-art explanatory power with high temporal stability, which significantly outperforms the industry-renowned expert-designed factors.
\end{itemize}

\section{Related Work}
\subsubsection{Style Factors by Human Experts}
The earliest Mean-Variance Model believed the risk should be measured while considering the return. And this risk is measured by the volatility of asset returns. The stocks with higher volatility of returns usually have a higher risk. It was the first time that mathematical statistics were introduced into portfolio theory, which built the foundation for modern finance. With the Mean-Variance Model proposed, scholars have gradually realized the importance of measuring portfolio risk. The CAPM divides the risk into systematic risk and unsystematic risk. And the unsystematic risk should not be compensated by high returns. The CAPM only uses one factor to measure the systematic risk. Due to the limited explanatory power of a single factor, the Multi-Factor Model approach was proposed. The Multi-Factor Model attributes the stock's return to several different style factors and uses those factors to estimate the covariance matrix. The main advantage of factor-based methods is to extract high-dimensional stock features into lower-dimensional factors so that the complexity of the problem will not change due to the number of stocks.
The most classic method in the Multi-Factor Models is Fama-French~\cite{8:fama1993common}. The Fama-French added value and size factors based on CAPM estimated the risk premium of the factors and tested the model's performance through time-series regression. In~\citeauthor{11:fama1973risk}[\citeyear{11:fama1973risk}], another cross-section regression factor model is proposed to determine the factor return at that cross-section regression. The~\citeauthor{19:fama2020comparing}[\citeyear{19:fama2020comparing}] pointed out that the cross-section method is better. Barra is a well-known cross-section regression multi-factor model. 

The above methods designed by the experts have high explanatory power, but the development is relatively slow due to the manual design requiring lots of human resources. It has spent almost decades from CAPM to Barra.

\subsubsection{Unsupervised Feature Extraction Methods}

Unsupervised machine learning has also been widely used in recent years. A simple and effective unsupervised method uses PCA to compress the high-dimensional stock data to the low-dimensional factor loading on the entire sequence. However, the factor loading generated by such a method does not change over time. As the stock distribution is unstable, the effect on the test set will gradually deteriorate. Therefore, \citeauthor{9:kelly2019characteristics}[\citeyear{9:kelly2019characteristics}] proposed a linear model of Instrumented PCA (IPCA), which estimates the factor load coefficient of style factor and assets from stock data. IPCA considers the asset characteristics as a variable, and the time changes can be included too. In addition, the decomposition process of Singular Value Decomposition (SVD)~\cite{30:golub1971singular} can also be implemented by AutoEncoder. However, this method does not use the temporal information of the stock data. Therefore, the FactorVAE~\cite{12:duan2022factorvae} is based on AutoEncoder adding Gated Recurrent Unit (GRU)~\cite{17:cho2014learning} to obtain the temporal information. But FactorVAE mainly focuses on the prediction ability of the model rather than the explanatory power.

\begin{figure*}[ht]
    \centering
    \includegraphics[width=6 in]{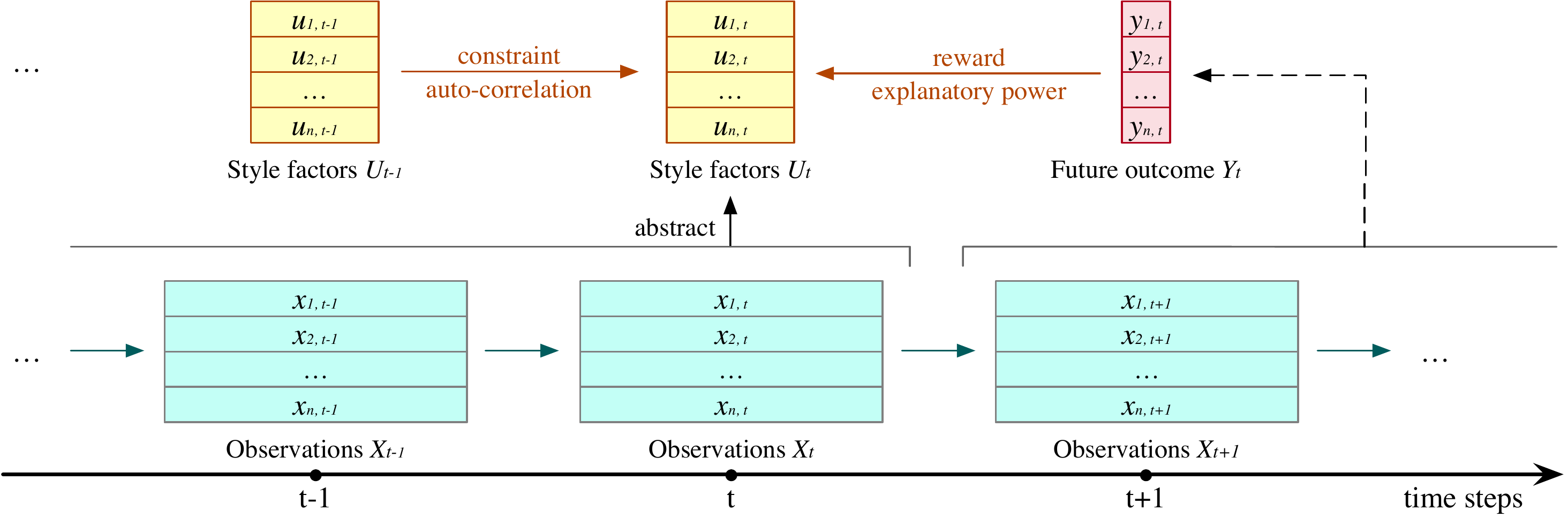}
    \caption{The overview of style factor extraction.\\}
    \label{fig2}
\end{figure*}

\subsubsection{Supervised Feature Extraction Method}
Supervised deep learning performs well in extracting complex non-linear features. It can also effectively use a supervised gradient to achieve the expected target. In DRM~\cite{7:lin2021deep}, style factors are used as input, and GRU and Graph Attention Network (GAT)~\cite{29:velivckovic2017graph} are added to discover the temporal and cross-section information. DRM designed a Multi-task method to smooth the output factors. DRM is a good attempt at deep learning in style factor extraction. However, Multi-task learning reduces training efficiency and hurts explanatory power because it is too concerned about long-term stability. Therefore, DRM has only relative increased by 2.0\% compared to the expert factors. In addition, DRM uses expert factors as input which leads to original information being lost, and its results largely depend on the quality of expert factors.

\section{Style Mining as a CMDP}
\newcommand{\obs}[1]{X_{{#1}}} 
\newcommand{\obsi}[1]{x_{{#1}}} 
\newcommand{\hidden}[1]{H_{{#1}}} 
\newcommand{\hiddeni}[1]{h_{{#1}}} %
\newcommand{\Length}[0]{L}
\newcommand{\state}[1]{S_{{#1}}} 
\newcommand{\statei}[1]{s_{{#1}}}
\newcommand{\statedim}[0]{I}
\newcommand{\sty}[1]{U_{{#1}}} 
\newcommand{\styi}[1]{u_{{#1}}}
\newcommand{\lab}[1]{Y_{{#1}}} 
\newcommand{\labi}[1]{y_{{#1}}} 

\newcommand{\constraincoef}[1]{\beta_{{#1}}}
\newcommand{\weight}[1]{w_{{#1}}}
\newcommand{\res}[1]{\epsilon_{{#1}}}
\newcommand{\reward}[1]{r_{{#1}}} 
\newcommand{\policy}[1]{\pi_{{#1}}} 
\newcommand{\stynum}[0]{K}
\newcommand{\stocknum}[0]{n}

\subsection{Notations}
As shown in Figure~\ref{fig2}, consider a time-series $\{\obs{t}\}_{t=1}^T$, where $\obs{t}=\{\obsi{i,t}\}_{i=1}^\stocknum$ has $\stocknum$ separate observations. For each $i=1,\dots,\stocknum$, the observation $\obsi{i,t}$ is a multi-dimensional vector. Further, let $\lab{t} = g(\obs{{\ge t}})$ be a future outcome for step $t$, which is the target label that we need to use the style factors to explain. And the $\lab{t} = \{ \labi{i,t}\}_{i=1}^\stocknum$ , where $\labi{i,t}$ is a value.

Such a setting is applicable in many areas. For example, in quantitative finance, $\obs{t}$ represents the market data at time-step $t$, $\stocknum$ is the number of stocks, $\obsi{i,t}$ represents the raw observation of the $i$-th stock at time $t$, and $\lab{t}$ can be the future returns or realized volatilities of these stocks.

Now, we need to find a set of style factors as a compressed representation of the historical data, denoted by $\sty{t} = \{\styi{i,t}\}^\stocknum_{i=1} = \policy{}(\obs{\leq t})$, where  $\styi{i,t}\in\mathbb{R}^{\stynum}$ is the $\stynum$-dimensional style factor for the $i$-th sequence. It is desired that these style factors can have strong explanatory power towards the target label $\lab{t}$ (i.e., a stock's styles can explain its future return), and also be stable across time as well (i.e., a stock's styles do not change in a short time). So we need to define the following two evaluation metrics.

\subsection{Evaluation Criteria of Style Factors}

\subsubsection{Explanatory power}
At each time step, suppose the true outcomes $\lab{t}$ are obtained, we can fit an explanatory model $\varphi(\sty{t})$ over the style factors, i.e., 
\begin{equation}
    \lab{t} = \hat{\lab{t}} + \res{t} = \varphi(\sty{t}) + \res{t},
    \label{eq:residual1}
\end{equation}
where $\res{t}=(\res{1,t}, \dots, \res{n, t})$ are the unexplained residuals. Usually, we can view the proportion of explained variance as explanatory power. Specifically, in this paper we use an averaged step-wise $R^2$ as the metric, i.e.,
\begin{equation}
\textrm{Expl}(\sty{1:T}) = \sum_{t=1}^T R^2_t=\sum_{t=1}^T\bigg[1-\frac{\sum^{\stocknum}_{i=1}\weight{i,t}\,\res{i,t}^2}{\sum^\stocknum _{i=1}\weight{i,t}\,\labi{i,t}^2}\bigg],
    \label{eq:R-square}
\end{equation}
where $\weight{i,t}$ is weight of the $i$-th sequence.
\subsubsection{Stability}
On the other hand, the style factors themselves as time-series should not fluctuate too much, which can be evaluated by the autocorrelation of the style series. Specifically, we use the lag-1 autocorrelation as follows:
\begin{equation}
    \textrm{AutoCorr}(\sty{1:T}) = \frac{1}{K}\sum_{k=1}^K \textrm{Corr}\big(\sty{1:T-1}^{(k)}, \sty{2:T}^{(k)}\big),
    \label{eq:autocor}
\end{equation}
where $\textrm{Corr}(\cdot, \cdot)$ is the Pearson correlation function.

As suggested by \citeauthor{28:menchero2011barra}[\citeyear{28:menchero2011barra}], style factors are considered high quality if they have autocorrelation coefficients above 0.9, otherwise too unstable to use. Therefore, in this paper we let $\textrm{AutoCorr}(\sty{1:T}) \geq 0.9$ as a constraint for stability.

\begin{figure*}[ht]
    \centering
    \includegraphics[width=7 in]{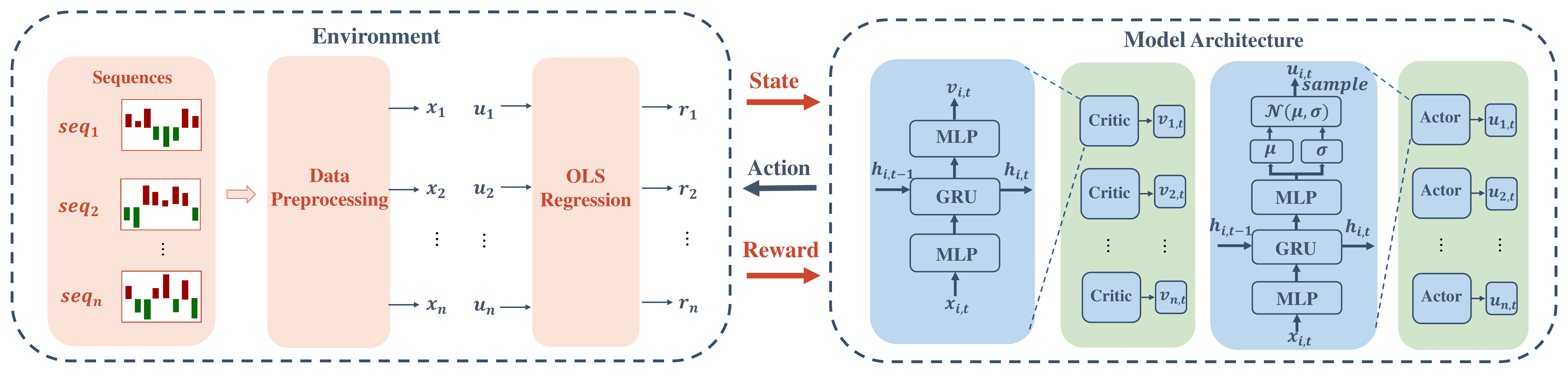}
    \caption{The overview design of Style Miner.}
    \label{fig:overview}
\end{figure*}

\subsection{Constrained Markov Decision Process}

We formulate the problem as a finite-horizon Constrained Markov Decision Process (CMDP)~\cite{32:altman1999constrained}, i.e.,
\begin{equation}
\begin{aligned}
    &\max_{\pi\in\Pi}\mathbb{E}_{\tau\sim\pi}[\mathcal{R}(\sty{1:T})], \\ 
    \subt\,&\mathbb{E}_{\tau\sim\pi}\left[\mathcal{C}(\sty{1:T})\right]\le d_0,
    \label{eq:CMDP}
\end{aligned}
\end{equation}
where $\mathcal{M}=\langle\state{},\sty{},\mathcal{P},\mathcal{R}, \mathcal{C}, d_0\rangle$, where $\state{t}$ is the state space which contain the history observation, $\sty{}$ is the action space, $\mathcal{P}$ is a transition function, $\mathcal{R}$ is the reward function, $\mathcal{C}$ is the cost function, and $d_0\in\mathbb{R}_{+}$ is the maximum allowed cost. With trajectory $\tau:=(\state{0},\sty{0},\state{1},\dots)$, where  $\sty{t}\sim\pi(\cdot|\state{t}),\state{t+1}\sim P(\cdot|\state{t},\sty{t})$, we use the explanatory power as the reward and stability as the constraint, i.e.,
\begin{equation}
\begin{aligned}
    &\mathcal{R}(\sty{1:T})={\textrm{Expl}(\sty{1:T})},\\
    &\mathcal{C}(\sty{1:T})={1-\textrm{AutoCorr}(\sty{1:T})}.
\end{aligned}
\end{equation}

\section{The Style Miner Algorithm}
There are two major challenges for the problem: 
\begin{enumerate}
    \item How to simplify the raw problem which has high problem complexity but sparse feedback.
    \item How to design an algorithm to efficiently balance the tradeoff between significance and stability.
\end{enumerate}

\subsection{Simplify the CMDP with Hidden States and Immediate Feedbacks}
From the previous section, we see that the problem has high-dimensional observations and actions but also has very few feedback signals. It is known that such a sparse reward problem can be hard to optimize. Therefore, in this section, we simplify the problem with three techniques: 1) the hidden state technique from sequential modelling to turn the high-dimensional sequential input into dense hidden representations, 2) representation for each sequence , and 3) the immediate rewards and costs to alleviate the sparse reward problem.

The details of the simplified CMDP are describe as following:

\subsubsection{Simplified states} There are two sources of complexity in the raw observations $\state{t}=\{\obs{\le t}\}$: 1) it includes $N$ multi-variable sequences, and $N$ can be large; 2) it includes $t$ historical time-steps, and $t$ can be large. Therefore, we suggest a simplified state setting with an independent encoder structure. We first decompose the overall state into individual sequences $\state{t}=\big\{\{\obsi{i,t}\}_t\big\}^n_{i=1}$ and assume an encoder that generates hidden states for each individual sequence $\hiddeni{i,t}=\textrm{Encoder}(\statei{i,t})$, so that $\tilde{s}_{i,t} = (\obsi{i,t}, \hiddeni{i,t-1})$.

Although not equivalent to the original problem definition, in this way the agent can take all the sequences as mini-batches and make decisions in parallel, which can empirically yields reasonable performance in practical problems.

\subsubsection{Immediate rewards} In the original setting, as the backward explanation and evaluation step takes all the sample sequences into account, every sequence shares the same cumulative return $\mathcal{R}(U_{1:T})$. To make full use of the minibatch-based policy iteration, we would like to distinguish the contribution of each individual sample sequence. Therefore, at each time step $t$, we decompose the overall R-squared $R_t^2$ into fine-grained immediate rewards for all the sequences. The $i$-th sequence gets the following reward:
\begin{equation}
    r_{i,t}=\frac{1}{\stocknum}-\frac{\weight{i,t}\styi{i,t}^2}{\sum^{\stocknum}_{i=1}\weight{i,t}\labi{i,t}^2}.
    \label{eq:reward decompose}
\end{equation}
So that we still have $r_t = R_t^2 = \sum_{i=1}^N r_{i,t}$. 


\subsubsection{Costs}  For each sequence $i$, the overall cost function is $\mathcal{C}(\styi{i,1:T}) = 1 - \textrm{AutoCorr}(\styi{i,1:T})$. As this cost can only be obtained after the episode is done, we define the immediate cost for each time-step $t$ as follows:
\begin{equation}
\begin{aligned}
\begin{split}
c(u_{i,t})= \left \{
\begin{array}{ll}
    \mathcal{C}(u_{i,1:T}),                    & \text{if}\quad t = T\\
    0,                                 & \text{otherwise}
\end{array}
\right.
.
\end{split}
\end{aligned}
\end{equation}

\subsection{Policy Optimization with Stability Constrain}
Based on the CMDP described above, we use Proximal Policy Optimization (PPO)~\cite{14:schulman2017proximal} to optimize a policy $\pi$.
PPO is an on-policy RL algorithm based on the trust region method, and it is applicable to continuous decision-making problems. It proposed a clipped version surrogate objective as follows:
\begin{small}
\begin{equation}
    L_{i,t}(\theta) =\min\left(\alpha_{i,t}(\theta)\hat{A}_{i,t},\textrm{clip}\left(\alpha_t(\theta),1-\epsilon,1+\epsilon\right)\hat{A}_{i,t}\right),
\end{equation}
\end{small}
where $\alpha_{i,t}(\theta)=\frac{\pi_{\theta}(\styi{i,t}|\statei{i,t})}{\pi_{old}(\styi{i,t}|\statei{i,t})}$, and the $\pi_{old}$ is the old policy before update, and $\hat{A}_{i,t}$ is the estimated advantage calculated as follow:

\begin{equation}
    \hat{A}_{i,t} = \sum^{k-1}_{j\ge 0}{\lambda^{j}\delta_{i,t+j}},
    \label{eq:adv}
\end{equation}
where $\delta_{i,t}=r_{i,t}+ V(\statei{i,t+1}) - V(\statei{i,t})$ and $\lambda$ is a discount coefficient to balance future errors. 


A common approach to solving a CMDP problem is using the Lagrange relaxation technique~\cite{32:altman1999constrained}. The Lagrange relaxation technique can turn the CMDP into an equivalent unconstrained problem by adding a penalty term, making the optimization process feasible.

However, in Style Miner, we found that our problem does not need to ensure that the constraints are met from beginning to end during the training process, but only the final result needs to meet the constraint. Therefore, we use a simpler heuristic method. Like the Lagrange approaches, it turns constraint into a regularization term with an adaptive multiplier $\constraincoef{}$. The multiplier has a linear increasing schedule, i.e, $\constraincoef{}' = \min(\constraincoef{} + \Delta \beta, \beta_{\max})$, until the constraint is met. Therefore, the surrogate loss can be formed as:
\begin{small}
\begin{equation}
\begin{split}
    L^{Actor}_{i,t} = &\hat{\mathbb{E}}_{i,t}\bigg[ \min\left(\alpha_{i,t}(\theta)\hat{A}_{i,t},\textrm{clip}\left(\alpha_{i,t}(\theta), 1-\epsilon,1+\epsilon\right)\hat{A}_{i,t}\right) \\
    & -\beta\max((c(\styi{i,t}) - d_0),0)\bigg].
    \label{eq:sm loss actor}
\end{split}
\end{equation}
\end{small}

After that, within the feasible solution range, Style Miner will optimize the explanatory power objective function in the unconstrained state. Therefore, this simple heuristic guarantees that Style Miner will finally achieve the constraints while optimizing the objective.


The loss function of the critic network is the mean-square-error between the output of critic network and the empirical value of cumulative future rewards $V_{i,t}^{target}$:
\begin{equation}
    L^{Critic}_{i,t} = \mathbb{E}_{i,t}\left[\left(V(\statei{i,t}) - V^{target}_{i,t}\right)^2\right],
    \label{eq:sm loss critic}
\end{equation}
where $V^{target}_{i,t}$ can be calculate as follows:
\begin{equation}
V^{target}_{i,t} = \sum_{t'>t}r_{i,t'}.
\label{eq:reward}
\end{equation}

\begin{algorithm}[h]
\caption{Training Procedure for Style Miner}
\textbf{Input}: inital policy parameters $\theta$, initial value function parameters $\phi$

\begin{algorithmic}[1] 
\FOR{each episode}
\FOR{each $t$}
\STATE Generate style factors $\styi{i,t} = \pi_{\theta}(\statei{i,t})$
\STATE Obtain the next state $\statei{i,t+1}$ and the reward $r_{i,t}$
\STATE Store the transition $\tau =(\statei{i,t}, \styi{i,t}, r_{i,t}, \statei{i,t+1})$
\ENDFOR
\STATE Compute cumulative future rewards $V^{target}_{i,t}$ according Eq.~\eqref{eq:reward}
\STATE Compute advantage estimates $\hat{A}_t$ according Eq.~\eqref{eq:adv} based on the current value function
\STATE Divide episode into data chunks 
\STATE Update the policy parameter $\theta$ to maximize the surrogate objective in Eq.~\eqref{eq:sm loss actor} for each data chunk via stochastic gradient ascent with Adam
\STATE Update the value function parameter $\phi$ according Eq.~\eqref{eq:sm loss critic}
\IF{the constraint is satisfy}
\STATE Deactivate the constraint penalty and stop update $\beta$
\ELSE{}
\IF{the constraint is satisfied before}
\STATE Activate the constraint penalty but not update $\beta$
\ELSE{}
\STATE Update $\beta$
\ENDIF{}
\ENDIF{}
\ENDFOR

\end{algorithmic}
\end{algorithm}

\subsubsection{The Style Miner Framework} 
The entire framework of Style Miner can be found in Figure~\ref{fig:overview}. Based on the actor-critic framework, we add the GRU both in Actor and Critic to obtain the hidden temporal information. We choose GRU due to it is simple and effective. The GRU architecture can be formulated as follows:
\begin{equation}
\begin{aligned}
    \hiddeni{i,t} &= \textrm{LayerNorm}(\textrm{GRU}(\obsi{i,t},\hiddeni{i,t-1})).
\end{aligned}
\end{equation}
The parameter $\theta$ of actor and critic between different sequences are shared.

During training, following the original PPO~\cite{14:schulman2017proximal}, the action is obtained by an action distribution, where the mean and variance from the policy output so the policy can achieve exploration.
Each episode has the same length, which is the length of the training dataset. After each episode terminates, the policy will update $k$ epochs. Different from PPO, we divide the entire episode into data trunks so we can train these data trunks in parallel and keep the original order in each data trunk. 

After the update, we will calculate the autocorrelation of the new policy on the validation dataset. Since it is time-consuming to calculate the autocorrelation of all stocks, we will sample dozens of stocks to get an approximate estimate. Once the autocorrelation meets the constraint, we will stop to increase the coefficient $\beta$.

The coefficient $\beta_{\max}$ is used to control the increased speed of the penalty term of constraint.
$d_0$ is the constrained term. 
Now we present our new algorithm, Style Miner, as shown in Algorithm 1, and all sequences are executed in parallel.

\section{Experiments}
Our experiments are based on style factor extraction on three datasets from the Chinese stock market with raw market data and expert-designed features. Through the experiments, we would like to answer three questions:

\textbf{Q1}: Whether Style Miner can generate style factors with both high explanatory power and high stability? 

\textbf{Q2}: Can we explicitly control the trade-off between them?

\textbf{Q3}: How about the performance of Style Miner with different features and different datasets?

\textbf{Q4}: How dose each part effect the performance of Style Miner? 

\subsection{Experiment settings}

\subsubsection{Data}
Our experiments are conducted on original raw data from the Chinese stock market. For the main experiments, we choose 13 commonly used raw features as the input:\{\emph{open}, \emph{high}, \emph{low}, \emph{close}, \emph{VWAP} (Volume-Weighted Average Price), \emph{volume}, \emph{money}, \emph{negMarketValue}, \emph{turnover rate}, \emph{amount of transactions}, \emph{PB}, \emph{PE}, \emph{percent-change}\}.
The dataset splits are based on datetimes ({2013/10/28 to 2017/12/29 for training, 2018/01/02 to 2018/12/28 for validation, and  2019/01/02 to 2019/12/30 for test}). 
Furthermore, we also train our algorithm on two datasets including CSI500, and CSI1000, to test the robustness and practicality of generated factors. 

\subsubsection{Compared Methods}
We organize and reproduce the existing style factor mining methods in detail, including expert factors~\cite{28:menchero2011barra}, unsupervised learning methods~\cite{5:wold1987principal,13:gu2021autoencoder,9:kelly2019characteristics} and supervised learning methods~\cite{7:lin2021deep}. In addition to several classical style factor generation methods currently available in the quantitative field, we have also implemented several reinforcement learning methods for comparison, including PPO~\cite{14:schulman2017proximal}, DDPG~\cite{20:lillicrap2015continuous} and TD3~\cite{td3}. A detail description of each baseline and hyperparameter of Style Miner can be found in Appendix.
We set the factor dimensions of all baselines equal to ten for a fair comparison.

\begin{figure*}[ht]
	\subfigure[]{
		\begin{minipage}[h]{0.48\linewidth}
			\centering
			\includegraphics[width=3.2in]{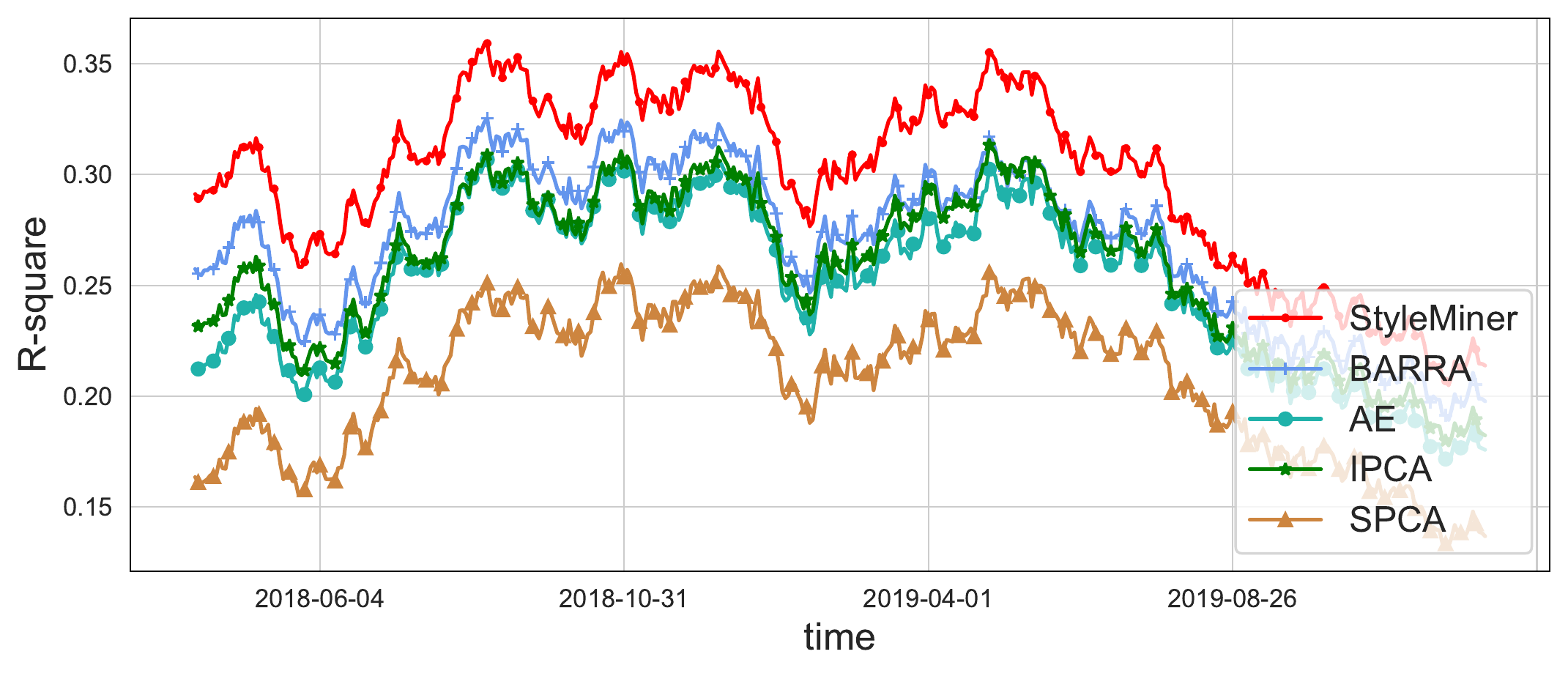}
			\label{fig:baseline compare}
		\end{minipage}
	}%
	\subfigure[]{
		\begin{minipage}[h]{0.48\linewidth}
			\centering
			\includegraphics[width=3.2in]{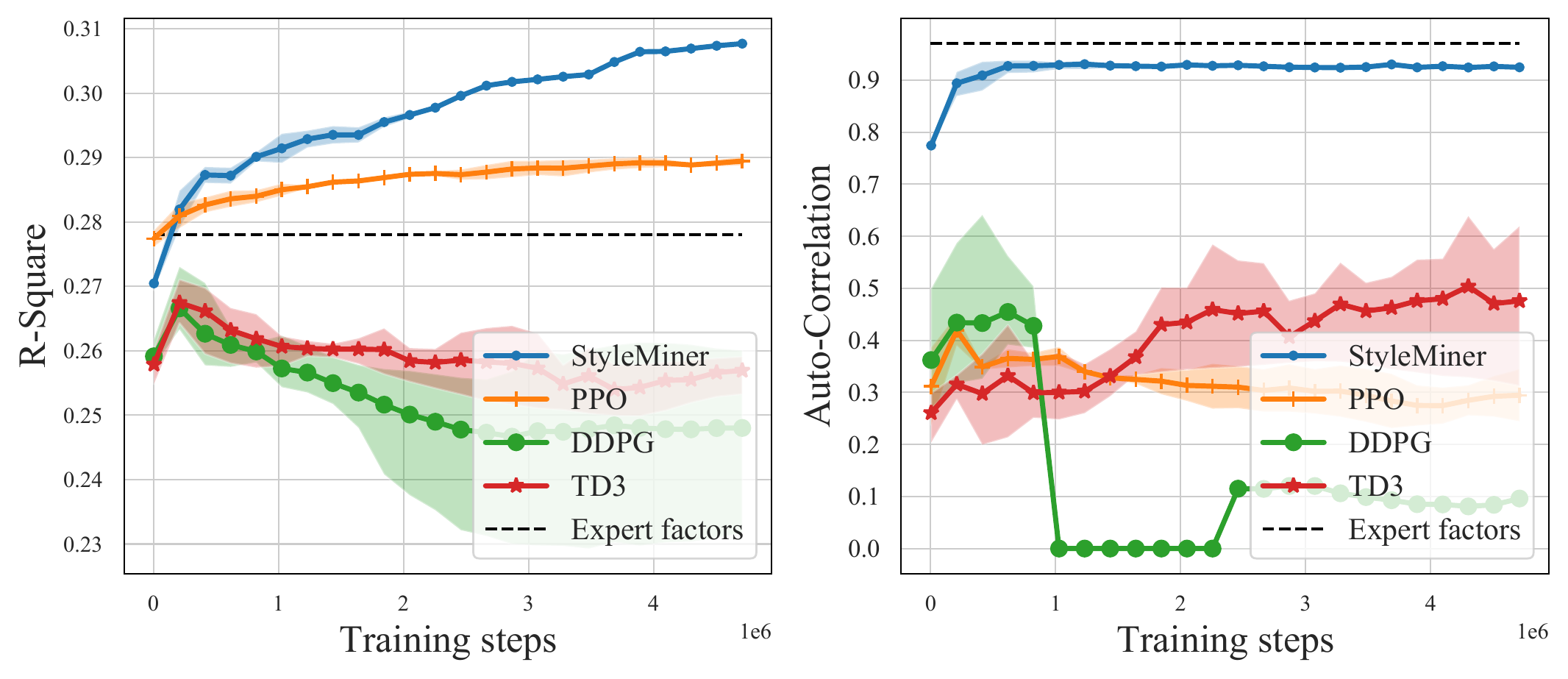}
			\label{fig:rl_method}
		\end{minipage}
	}%
 
	\subfigure[]{
		\begin{minipage}[h]{0.48\linewidth}
			\centering
			\includegraphics[width=3.2in]{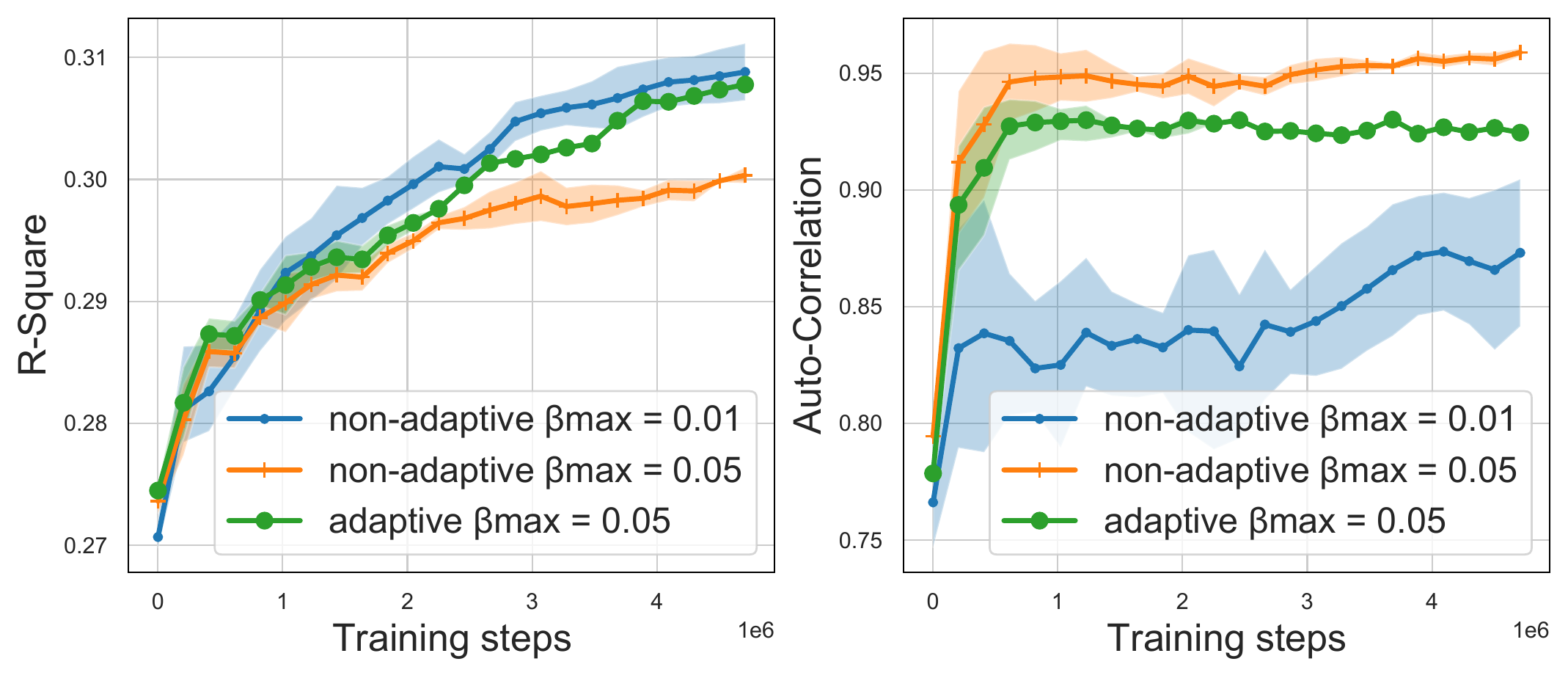}
			\label{fig:coef}
		\end{minipage}
	}%
	\subfigure[]{
		\begin{minipage}[h]{0.48\linewidth}
			\centering
			\includegraphics[width=3.2in]{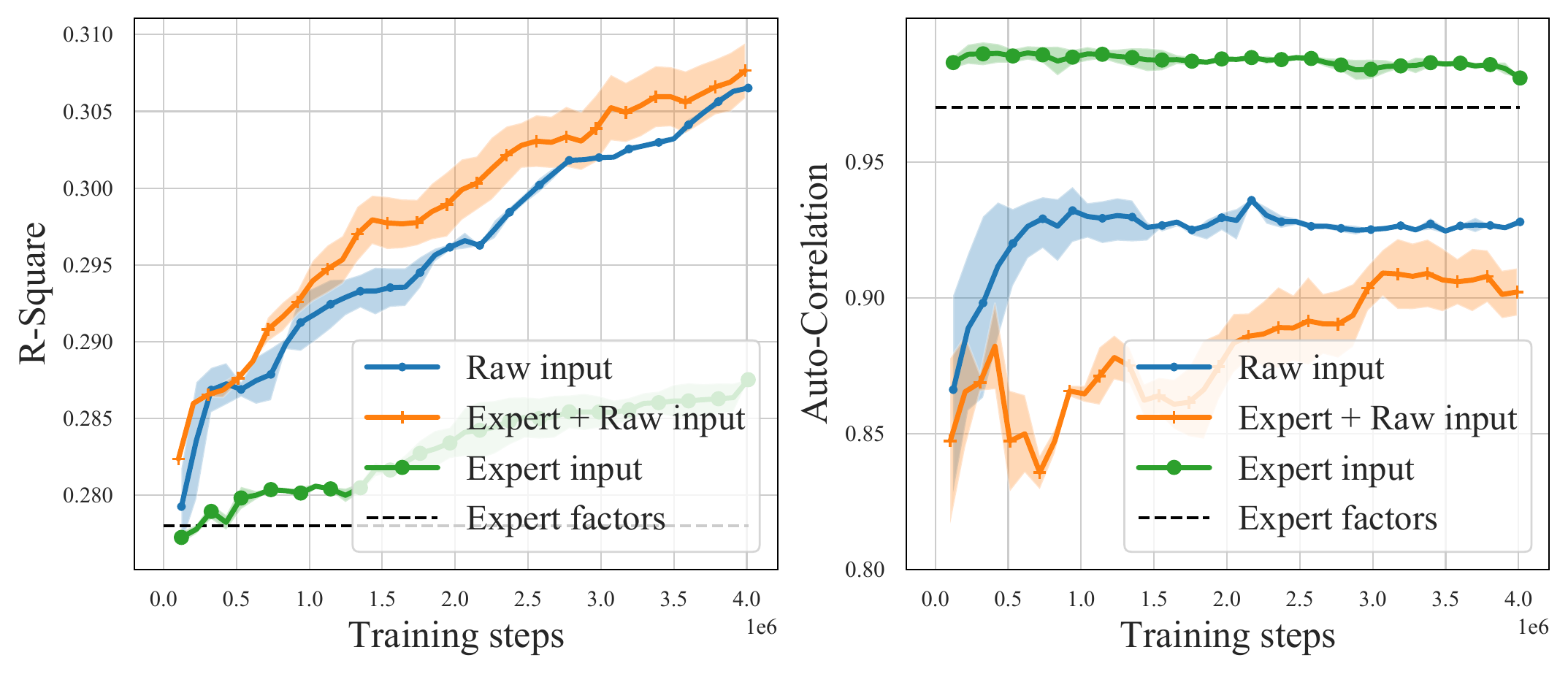}
			\label{fig:features}
		\end{minipage}
	}%
	\caption{Experiments results. The solid lines are averaged cumulative regrets over three random seeds, and the shaded areas stand for the standard deviations. (a) The R-square of different algorithms on the test set (60 days moving averages). (b) The comparison between different RL methods on valid dataset. (c) The comparison between different $\beta_{\max}$. (d) The comparison between different input features.}
	\label{fig4: experiments}
\end{figure*}


\subsubsection{Data Preprocessing}
The price columns (\emph{open}, \emph{high}, \emph{low}, \emph{close}, and \emph{VWAP}) have similar values, so we replace \emph{high}, \emph{low}, \emph{close}, and \emph{VWAP} with the percent changes relative to the open price. All feature columns are standardized with Z-scores. In addition, we used forward fill to handle the missing values.

\subsubsection{Evalutation Metrics} We choose three metrics as follow:
\begin{itemize}
    \item $\mathbf{R^2}$ is used to measures the explanatory power of model. The calculation of $R^2$ is defined as Eq.~\eqref{eq:R-square}.
    \item \textbf{Avg. T-value} is the average absolute t-statistic of all style factors. The T-value shows the significant of the style factors.
    \item \textbf{AutoCorr} is the average of the autocorrelation coefficients of all factors over all stock series. The calculation of autocorrelation is defined as Eq.~\eqref{eq:autocor}.
\end{itemize}
The above three metrics are all the higher, the better.

\subsection{Experiment Results}
\subsubsection{Main Results}
To answer \textbf{Q1}, we first compared several well-established baseline models with Style Miner on three major metrics: $R^2$, Average T-value of all factors, and AutoCorr. The environment is based on the full stocks with its raw data, and the results are calculated on the test dataset.
\begin{table}[t]
\centering
\setlength{\tabcolsep}{1.1mm}{
\small{
\begin{tabular}{cccc}
\toprule
Algorithms          & $R^2$     & Avg. T-value & AutoCorr\\
\midrule
ExpertFactors       & 24.8\%    & 2.89                  & \bf{0.97}\\
PCA                & 19.1\%    & 0.85                  & -\\
IPCA                & 23.8\%    & 4.11         & 0.29\\
AutoEncoder         & 23.1\%    & 3.32                  & 0.21\\
PPO                 & 25.4\%    & 2.16                  & 0.39\\
DDPG                & 24.5\%    & 2.71                  & 0.22\\
TD3                 & 24.7\%    & 2.57                  & 0.33\\
\midrule
StyleMiner & \textbf{27.4\%} & \textbf{4.43} & \bf{0.92} \\
\midrule
StyleMiner (post-EMA)  & 26.5\% & 3.81 & \bf{0.98}\\
StyleMiner (Expert input) & 25.6\% & 2.71 & \textbf{0.99} \\
StyleMiner (Expert+Raw input) & \textbf{27.6}\% & 4.38 & \textbf{0.93} \\
\bottomrule
\end{tabular}
}}
\caption{The performance of the compared algorithms.}
\label{baseline tabel}
\end{table}
The results are shown in Table~\ref{baseline tabel}. Under the premise of using raw data as input, Style Miner can achieve a 10.4\% relative gain in $R^2$ than the expert factors with the AutoCorr above 0.92, which satisfies the stability requirement. The PCA method has no Autocorr since it is fixed values on the entire series.

For the situation that needs to achieve a higher AutoCorr, we also provide an Exponential Moving Average (EMA) version of Style Miner, i.e., using a simple weighted average over the generated sequence of style factors. The EMA can be formulated as $
\sty{t} = \rho \sty{t} +(1-\rho) \sty{t-1}$, here we set $\rho = 0.25$. Since the style factors generated by Style Miner already have high stability, it can achieve an extremely high AutoCorr with less $R^2$ lost. The final result of the EMA version of Style Miner with daily features can still relatively exceed 6.4\% $R^2$ than the expert factors with the same AutoCorr. 
And the $R^2$ of each day on the validation and test dataset is shown in Figure~\ref{fig:baseline compare}. It can be seen that on the entire validation and test set, Style Miner can achieve a far better explanatory power than the current public methods.

To answer the \textbf{Q2}, we compared the performance of Style Miner without adaptive penalty, which means the penalty term will keep affecting the policy even if the constraint is met. As shown in Fig.~\ref{fig:coef}, Style Miner with adaptive $\beta_{\max}=0.05$ can achieve a nearly $R^2$ to non-adaptive $\beta_{\max}=0.01$ and still keep a high autocorrelation.

\subsubsection{Comparison with Supervised Learning Method}
We reproduce Deep Risk Model(DRM), which is a supervised learning method, and make detailed comparisons with Style Miner in Table.~\ref{drm}. The parameter H is the horizon of multi-task learning in DRM. The larger the H, the smoother the model output. The experimental results are shown in Table~\ref{drm}. 

\begin{table}[htbp]
\centering
\setlength{\tabcolsep}{1.1mm}{
\small{
\begin{tabular}{cccc}
\toprule
Algorithms          & $R^2$     & Avg. T-value & AutoCorr\\
\midrule
ExpertFactors       & 24.8\%    & 2.89    & \bf{0.97}\\
\midrule
DRM (H =  ~1, Raw) & 26.4\%  & 4.36 & 0.20\\
DRM (H = 20, Raw) & 24.6\%    &  3.49 & 0.57\\
DRM (H = ~1, Expert) & 25.0\%   & 2.64  & \textbf{0.98} \\
DRM (H = 20, Expert) & 25.0\%  & 2.74  & \textbf{0.99}\\
DRM (H =  ~1, Raw + Expert) & 27.4\%   & 3.88 & 0.56 \\
DRM (H = 20, Raw + Expert) & 25.6\%  & 3.48 & 0.70\\
\midrule
StyleMiner (Raw)& \textbf{27.4\%} & \textbf{4.43} & \bf{0.92} \\
StyleMiner (Raw + EMA)  & 26.5\% & 3.81 & \bf{0.98}\\
StyleMiner (Expert) & 25.6\% & 2.71 & \textbf{0.99} \\
StyleMiner (Expert+Raw) & \textbf{27.6}\% & 4.38 & \textbf{0.93} \\
\bottomrule
\end{tabular}
}}
\caption{The performance of the compared algorithms.}
\label{drm}
\end{table}

By comparing different types of input data, we can find that DRM has higher requirements for input data, and in the case of raw data input, DRM cannot obtain high stability. So the multi-task learning method in DRM can not promise stability and will lead to the problems of slow training speed and poor explanatory power. In addition, under the same input, Style Miner can obtain explanatory power and stability far exceeding DRM.

\subsubsection{Comparison with Reinforcement Learning Methods} 
The training curves of all RL-based methods are shown in \figurename~\ref{fig:rl_method}. Although the other reinforcement learning methods can mine style factors with close explanatory power to expert factors, the generated style factors are unstable.



\subsubsection{Different Input Features and Datasets}
We further investigate the performance of Style Miner with different features (\textbf{Q3}). As shown in Figure~\ref{fig:features}, when using expert factors as input, the model will achieve a high autocorrelation due to the input being quite stable. However, the explanatory power is also limited due to the smooth input. By using both raw data and expert factors as input, Style Miner can achieve a higher $R^2$ with autocorrelation above 0.9. Therefore, we can conclude that with smoother input features, Style Miner can obtain a higher interpretation ratio while maintaining higher stability. Additional input features can further help improve the explanatory power of the model. 

We also test two datasets from the Chinese stock market, including CSI500 and CSI1000, which contain 500 and 1000 stocks. We show CSI500 in Table~\ref{csi500}, CSI1000 and the time cost of Style Miner and PPO on three datasets can be found in Appendix. The results show that, since the scale of stocks is decreasing, the output of other models has also become more stable but still does not meet the requirements (Above 0.9).

\begin{table}[h]
\centering
\setlength{\tabcolsep}{5.1mm}{
\small{
\begin{tabular}{lcc}
\toprule
Algorithms          & $R^2$    & AutoCorr\\
\midrule
StyleMiner & \textbf{31.1\%}  & \bf{0.92} \\
PPO  & 28.7\% & 0.75\\
DDPG & 28.0\% & 0.85\\
TD3 & 28.2\% & 0.78\\
Expert & 27.7\% & \bf{0.97}\\
\bottomrule
\end{tabular}
}}
\caption{Results on CSI500.}
\label{csi500}
\end{table}

\subsubsection{Ablation Study}
To answer \textbf{Q4}, we did an ablation study on each technical part of StyleMiner. The results on all stocks are shown in Table~\ref{ablation}.
The ablation results show that each part is essential, and we can get the below analysis: GRU makes the model can obtain history information, otherwise it cannot achieve high smoothness; Constraint can balance the tradeoff since we do not need a too high smoothness; Penalty can promise high AutoCorr; Reward decomposed methods can identify the contribution of each stock, therefore model can achieve high $R^2$.

\begin{table}[t]
\centering
\setlength{\tabcolsep}{1.1mm}{
\small{
\begin{tabular}{lcc}
\toprule
Algorithms          & $R^2$    & AutoCorr\\
\midrule
StyleMiner(SM) & \textbf{27.4\%}  & \bf{0.92} \\
SM w/o GRU   & 26.0\%(-5.2\%) & 0.51 (-45\%)\\
SM w/o Constraint   & 26.2\%(-4.4\%) & \bf{0.97(+4\%)}\\
SM w/o AutoCorr Penalty   & \textbf{27.6}\%(+0.7\%) & 0.73(-21\%)\\
SM w/o Reward Decomposed   & 24.3\%(-11\%) & 0.81(-12\%)\\

\bottomrule
\end{tabular}
}}
\caption{The ablation study for each part.}
\label{ablation}
\end{table}

\section{Conclusion}
In this paper, we proposed a novel reinforcement learning algorithm for style factor extraction called Style Miner. To solve the challenge in style factor extraction, we formulate the style mining as a CMDP, design fine-grained immediate rewards and costs, and use a Lagrangian heuristic to adaptively balance between them. The experiment results show that Style Miner significantly outperforms other baselines.






\bibliographystyle{named}
\bibliography{ijcai23}

\begin{thebibliography}{}

\bibitem[\protect\citeauthoryear{Altman}{1999}]{32:altman1999constrained}
Eitan Altman.
\newblock {\em Constrained Markov decision processes: stochastic modeling}.
\newblock Routledge, 1999.

\bibitem[\protect\citeauthoryear{Bahdanau \bgroup \em et al.\egroup
  }{2015}]{35:bahdanau2014neural}
Dzmitry Bahdanau, Kyunghyun Cho, and Yoshua Bengio.
\newblock Neural machine translation by jointly learning to align and
  translate.
\newblock In Yoshua Bengio and Yann LeCun, editors, {\em 3rd International
  Conference on Learning Representations, {ICLR} 2015, San Diego, CA, USA, May
  7-9, 2015, Conference Track Proceedings}, 2015.

\bibitem[\protect\citeauthoryear{Bindoff \bgroup \em et al.\egroup
  }{2013}]{25:bindoff2013detection}
Nathaniel~L Bindoff, Peter~A Stott, Krishna~Mirle AchutaRao, Myles~R Allen,
  Nathan Gillett, David Gutzler, Kabumbwe Hansingo, Gabriele Hegerl, Yongyun
  Hu, Suman Jain, et~al.
\newblock Detection and attribution of climate change: from global to regional.
\newblock 2013.

\bibitem[\protect\citeauthoryear{Cho \bgroup \em et al.\egroup
  }{2014}]{17:cho2014learning}
Kyunghyun Cho, Bart van Merri{\"e}nboer, Caglar Gulcehre, Dzmitry Bahdanau,
  Fethi Bougares, Holger Schwenk, and Yoshua Bengio.
\newblock Learning phrase representations using {RNN} encoder{--}decoder for
  statistical machine translation.
\newblock In {\em Proceedings of the 2014 Conference on Empirical Methods in
  Natural Language Processing ({EMNLP})}, pages 1724--1734, Doha, Qatar, 2014.
  Association for Computational Linguistics.

\bibitem[\protect\citeauthoryear{Duan \bgroup \em et al.\egroup
  }{2022}]{12:duan2022factorvae}
Yitong Duan, Lei Wang, Qizhong Zhang, and Jian Li.
\newblock Factorvae: A probabilistic dynamic factor model based on variational
  autoencoder for predicting cross-sectional stock returns.
\newblock 2022.

\bibitem[\protect\citeauthoryear{Fama and French}{1993}]{8:fama1993common}
Eugene~F Fama and Kenneth~R French.
\newblock Common risk factors in the returns on stocks and bonds.
\newblock {\em Journal of financial economics}, 33(1):3--56, 1993.

\bibitem[\protect\citeauthoryear{Fama and French}{2020}]{19:fama2020comparing}
Eugene~F Fama and Kenneth~R French.
\newblock Comparing cross-section and time-series factor models.
\newblock {\em The Review of Financial Studies}, 33(5):1891--1926, 2020.

\bibitem[\protect\citeauthoryear{Fama and MacBeth}{1973}]{11:fama1973risk}
Eugene~F Fama and James~D MacBeth.
\newblock Risk, return, and equilibrium: Empirical tests.
\newblock {\em Journal of political economy}, 81(3):607--636, 1973.

\bibitem[\protect\citeauthoryear{Fujimoto \bgroup \em et al.\egroup
  }{2018}]{td3}
Scott Fujimoto, Herke van Hoof, and David Meger.
\newblock Addressing function approximation error in actor-critic methods.
\newblock In Jennifer~G. Dy and Andreas Krause, editors, {\em Proceedings of
  the 35th International Conference on Machine Learning, {ICML} 2018,
  Stockholmsm{\"{a}}ssan, Stockholm, Sweden, July 10-15, 2018}, volume~80 of
  {\em Proceedings of Machine Learning Research}, pages 1582--1591. {PMLR},
  2018.

\bibitem[\protect\citeauthoryear{Golub and
  Reinsch}{1971}]{30:golub1971singular}
Gene~H Golub and Christian Reinsch.
\newblock Singular value decomposition and least squares solutions.
\newblock In {\em Linear algebra}, pages 134--151. Springer, 1971.

\bibitem[\protect\citeauthoryear{Gu \bgroup \em et al.\egroup
  }{2021}]{13:gu2021autoencoder}
Shihao Gu, Bryan Kelly, and Dacheng Xiu.
\newblock Autoencoder asset pricing models.
\newblock {\em Journal of Econometrics}, 222(1):429--450, 2021.

\bibitem[\protect\citeauthoryear{Guy.}{1975}]{2:1975The}
Rosenberg~James Guy.
\newblock The prediction of systematic risk.
\newblock {\em Research Program in Finance Working Papers}, 1975.

\bibitem[\protect\citeauthoryear{Hegerl and Zwiers}{2011}]{23:hegerl2011use}
Gabriele Hegerl and Francis Zwiers.
\newblock Use of models in detection and attribution of climate change.
\newblock {\em Wiley interdisciplinary reviews: climate change}, 2(4):570--591,
  2011.

\bibitem[\protect\citeauthoryear{Ilmanen}{2012}]{21:ilmanen2012understanding}
Antti Ilmanen.
\newblock Understanding expected returns.
\newblock In {\em CFA Institute Conference Proceedings Quarterly}, 2012.

\bibitem[\protect\citeauthoryear{Kelly \bgroup \em et al.\egroup
  }{2019}]{9:kelly2019characteristics}
Bryan~T Kelly, Seth Pruitt, and Yinan Su.
\newblock Characteristics are covariances: A unified model of risk and return.
\newblock {\em Journal of Financial Economics}, 134(3):501--524, 2019.

\bibitem[\protect\citeauthoryear{Lillicrap \bgroup \em et al.\egroup
  }{2016}]{20:lillicrap2015continuous}
Timothy~P. Lillicrap, Jonathan~J. Hunt, Alexander Pritzel, Nicolas Heess, Tom
  Erez, Yuval Tassa, David Silver, and Daan Wierstra.
\newblock Continuous control with deep reinforcement learning.
\newblock In Yoshua Bengio and Yann LeCun, editors, {\em 4th International
  Conference on Learning Representations, {ICLR} 2016, San Juan, Puerto Rico,
  May 2-4, 2016, Conference Track Proceedings}, 2016.

\bibitem[\protect\citeauthoryear{Lin \bgroup \em et al.\egroup
  }{2021}]{7:lin2021deep}
Hengxu Lin, Dong Zhou, Weiqing Liu, and Jiang Bian.
\newblock Deep risk model: a deep learning solution for mining latent risk
  factors to improve covariance matrix estimation.
\newblock In {\em Proceedings of the Second ACM International Conference on AI
  in Finance}, pages 1--8, 2021.

\bibitem[\protect\citeauthoryear{Menchero \bgroup \em et al.\egroup
  }{2011}]{28:menchero2011barra}
Jose Menchero, D~Orr, and Jun Wang.
\newblock The barra us equity model (use4), methodology notes.
\newblock {\em English, MSCI May}, 2011.

\bibitem[\protect\citeauthoryear{Pan \bgroup \em et al.\egroup }{2022}]{lcad}
Feiyang Pan, Tongzhe Zhang, Ling Luo, Jia He, and Shuoling Liu.
\newblock Learn continuously, act discretely: Hybrid action-space reinforcement
  learning for optimal execution, 2022.

\bibitem[\protect\citeauthoryear{Ping~Tian and
  others}{2013}]{34:ping2013review}
Dong Ping~Tian et~al.
\newblock A review on image feature extraction and representation techniques.
\newblock {\em International Journal of Multimedia and Ubiquitous Engineering},
  8(4):385--396, 2013.

\bibitem[\protect\citeauthoryear{Schulman \bgroup \em et al.\egroup
  }{2017}]{14:schulman2017proximal}
John Schulman, Filip Wolski, Prafulla Dhariwal, Alec Radford, and Oleg Klimov.
\newblock Proximal policy optimization algorithms.
\newblock {\em arXiv preprint arXiv:1707.06347}, 2017.

\bibitem[\protect\citeauthoryear{Sharpe}{1964}]{4:sharpe1964capital}
William~F Sharpe.
\newblock Capital asset prices: A theory of market equilibrium under conditions
  of risk.
\newblock {\em The journal of finance}, 19(3):425--442, 1964.

\bibitem[\protect\citeauthoryear{Sheikh}{1996}]{3:sheikh1996barra}
Aamir Sheikh.
\newblock Barra's risk models.
\newblock {\em Barra Research Insights}, pages 1--24, 1996.

\bibitem[\protect\citeauthoryear{Velickovic \bgroup \em et al.\egroup
  }{2018}]{29:velivckovic2017graph}
Petar Velickovic, Guillem Cucurull, Arantxa Casanova, Adriana Romero, Pietro
  Li{\`{o}}, and Yoshua Bengio.
\newblock Graph attention networks.
\newblock In {\em 6th International Conference on Learning Representations,
  {ICLR} 2018, Vancouver, BC, Canada, April 30 - May 3, 2018, Conference Track
  Proceedings}. OpenReview.net, 2018.

\bibitem[\protect\citeauthoryear{Wold \bgroup \em et al.\egroup
  }{1987}]{5:wold1987principal}
Svante Wold, Kim Esbensen, and Paul Geladi.
\newblock Principal component analysis.
\newblock {\em Chemometrics and intelligent laboratory systems}, 2(1-3):37--52,
  1987.

\end{thebibliography}

\newpage
\appendix
\section{Detail description of baselines}
In this section, we  want to describe all our baseline methods.
\begin{itemize}
    \item \textbf{Expert Factors}~\cite{28:menchero2011barra} is  ten style factor including Size, Beta, Momentum, Residual Volatility, Non-linear Size, Book-to-Price, Liquidity, Earnings Yield, Growth and Leverage. It shows the best performance of style factors designed by experts.
    \item \textbf{PCA}~\cite{5:wold1987principal} is the unsupervised learning method that uses PCA to extract ten factors with the largest eigenvalues from the stock data series.
    \item \textbf{Auto Encoder} \cite{13:gu2021autoencoder} is a latent factor conditional asset pricing model which use Auto Encoder to obtain non-linear and time-varying factor exposures.
    \item \textbf{IPCA}~\cite{9:kelly2019characteristics} introduces company characteristics as the variable of factor exposure and excess return, which solves the problem that factor loading does not change with time in SPCA.
    \item \textbf{DDPG}~\cite{20:lillicrap2015continuous} is an off-policy reinforcement learning method that uses the Actor-Critic architecture and adopts a dual network structure.
    \item \textbf{PPO}~\cite{14:schulman2017proximal} is a policy-based reinforcement learning method. PPO proposed a clipped objective function that can be updated in small batches with multiple training steps.
    \item \textbf{TD3}~\cite{td3} can be considered an improved version of DDPG, it combines ideas from several different approaches to stabilize the learning process. 
    \item \textbf{DRM}~\cite{7:lin2021deep} is a supervised method use GRU and GAT to extract style factors.

\end{itemize}

\section{Hyperparameters}
The detail hyperparameters of Style Miner are shown in Table~ \ref{table:hyperparameters}.
\begin{table}[htbp]
\centering
\setlength{\tabcolsep}{5 mm}{
\begin{tabular}{ll}
\hline
 \textbf{Description}                                        & \textbf{Value} \\ \hline
             Type of optimizer& Adam        \\
             Learning rate of Actor                                               & 0.00001         \\
                                Learning rate of Critic                                              &0.00001         \\
                 Training epochs of each update                 & 10            \\
                 Batch size                                                  & 16             \\
                 Capacity of replay buffer (in episodes)                     & 1024           \\
                  Length of each data trunk   & 20   \\ 
                   The hidden state size of GRU   & 256   \\ 
                   Maximum value of action   & 5   \\ 
                   Activate function   & Tanh   \\ 
                   $\lambda$ in GAE   & 0.95   \\ 
                   Constrained $d_0$ & 0.075 \\
                                   \hline
\end{tabular}}
\caption{Hyperparameter settings in Style Miner.}
\label{table:hyperparameters}
\end{table}

\section{Additional Experiments}

\subsection{External Dataset}
We show the performance of Reinforcement Learning methods and expert factor on CSI1000 in Table~\ref{csi1000}.
\begin{table}[htbp]
\centering
\setlength{\tabcolsep}{5.1mm}{
\small{
\begin{tabular}{lcc}
\toprule
Algorithms          & $R^2$    & AutoCorr\\
\midrule
StyleMiner & \textbf{31.1\%}  & \bf{0.91} \\
PPO  & 27.3\% & 0.55\\
DDPG & 25.6\% & 0.81\\
TD3 & 26.1\% & 0.69\\
Expert & 25.3\% & \bf{0.96}\\
\bottomrule
\end{tabular}
}}
\caption{Results on CSI1000.}
\label{csi1000}
\end{table}

\subsection{Cost Analyse}
In this section, we want to show the comparison of cost time between StyleMiner and PPO on each dataset. 
\begin{figure}[htbp]
    \centering
    \includegraphics[width=3 in]{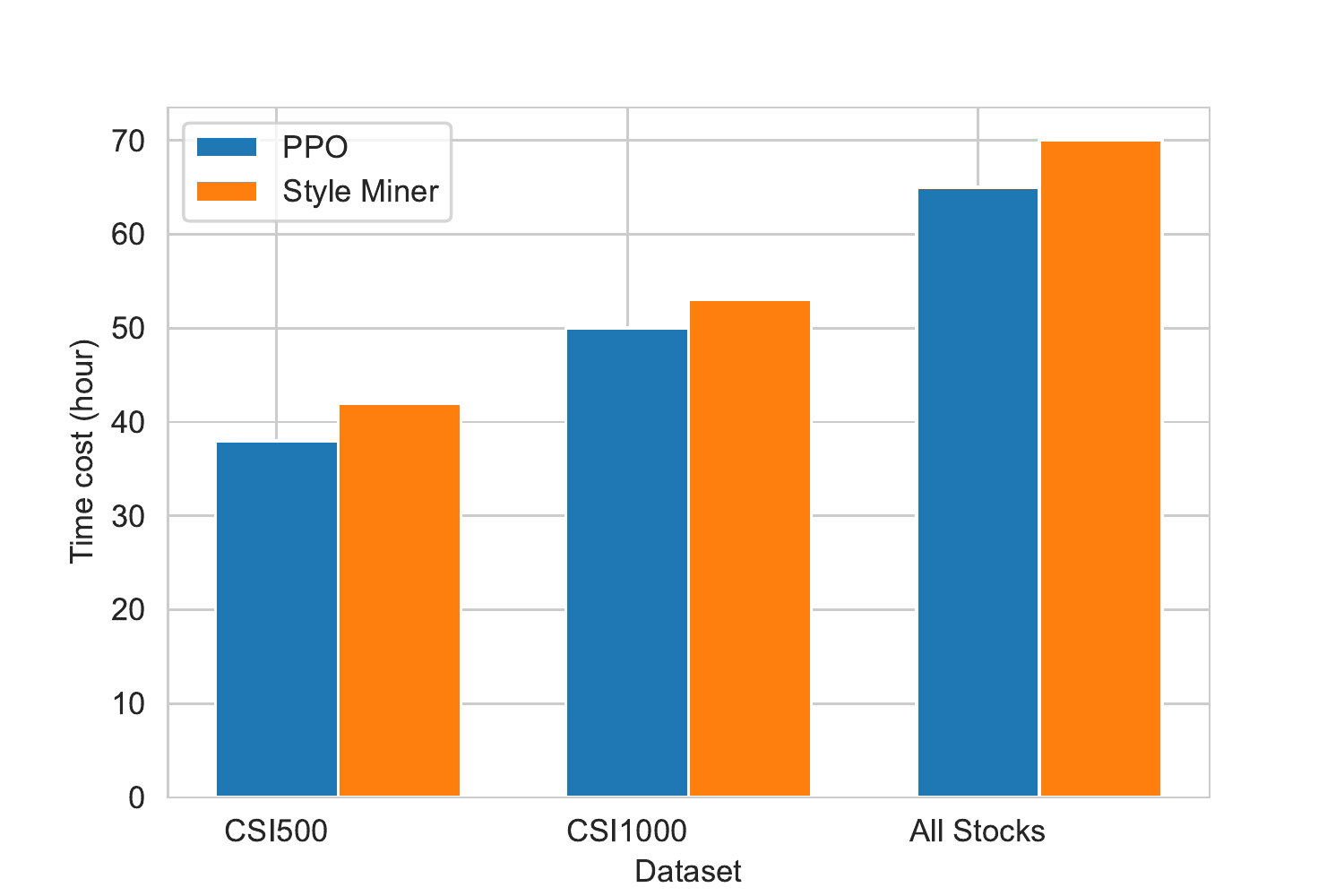}
    \caption{Cost time comparison.}
    \label{fig:cost time}
\end{figure}

As shown in Figure~\ref{fig:cost time}, the more stocks in the dataset, the more time cost. But the differences between PPO and StyleMiner are always small.

\subsection{Factor Visualization}
\begin{figure}[htbp]
    \centering
    \includegraphics[width=3.2 in]{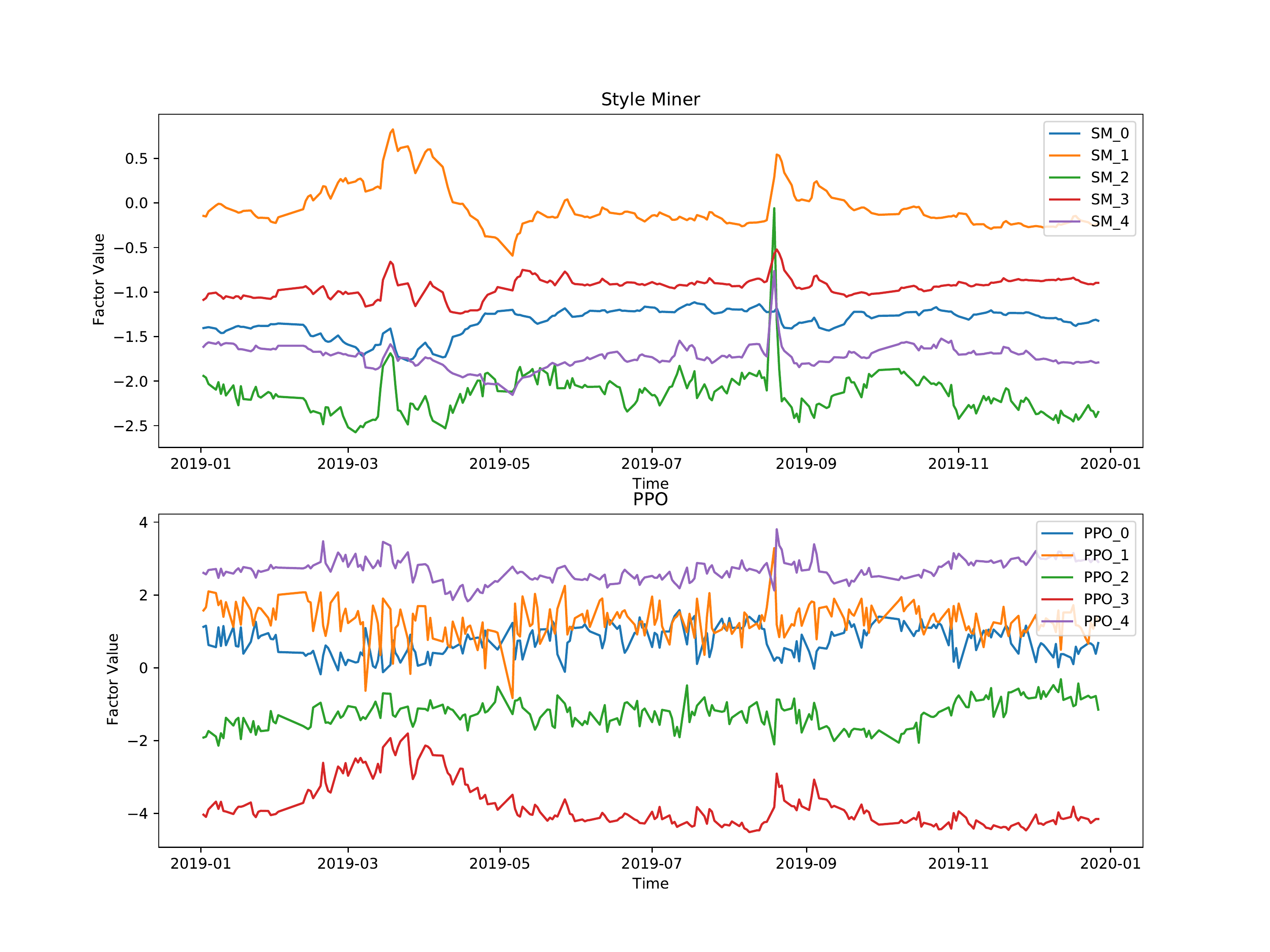}
    \caption{Factor value.}
    \label{fig:factors}
\end{figure}

In order to display the style factors more intuitively, we visualized first five factors values on a random stock generated by Style Miner and PPO according to the original data for the test set. The Figure~\ref{fig:factors} shows that Style Miner has smoother output all the time.
\end{document}